\documentclass[a4paper,conference]{IEEEtran}
\IEEEoverridecommandlockouts

\usepackage{cite}
\usepackage{amsmath,amssymb}
\usepackage{graphicx}
\usepackage{booktabs}
\usepackage{array}

\newif\ifanonymoussubmission
\anonymoussubmissionfalse

\begin{document}

\title{TRACT: Temporally Routed Action Chunks with Chronological Phase Authority for Contact-Rich Manipulation}

\ifanonymoussubmission
\author{\IEEEauthorblockN{Anonymous Authors}}
\else
\author{
\IEEEauthorblockN{Jiahao Liu, Kento Kawaharazuka, Tasuku Makabe, Kei Okada}
\IEEEauthorblockA{
\textit{Department of Mechano-Informatics} \\
\textit{Graduate School of Information Science and Technology, The University of Tokyo} \\
Tokyo, Japan \\
\{liu, kawaharazuka, makabe, k-okada\}@jsk.imi.i.u-tokyo.ac.jp
}
}
\fi

\IEEEaftertitletext{%
\begin{center}
\centering
\includegraphics[width=\textwidth]{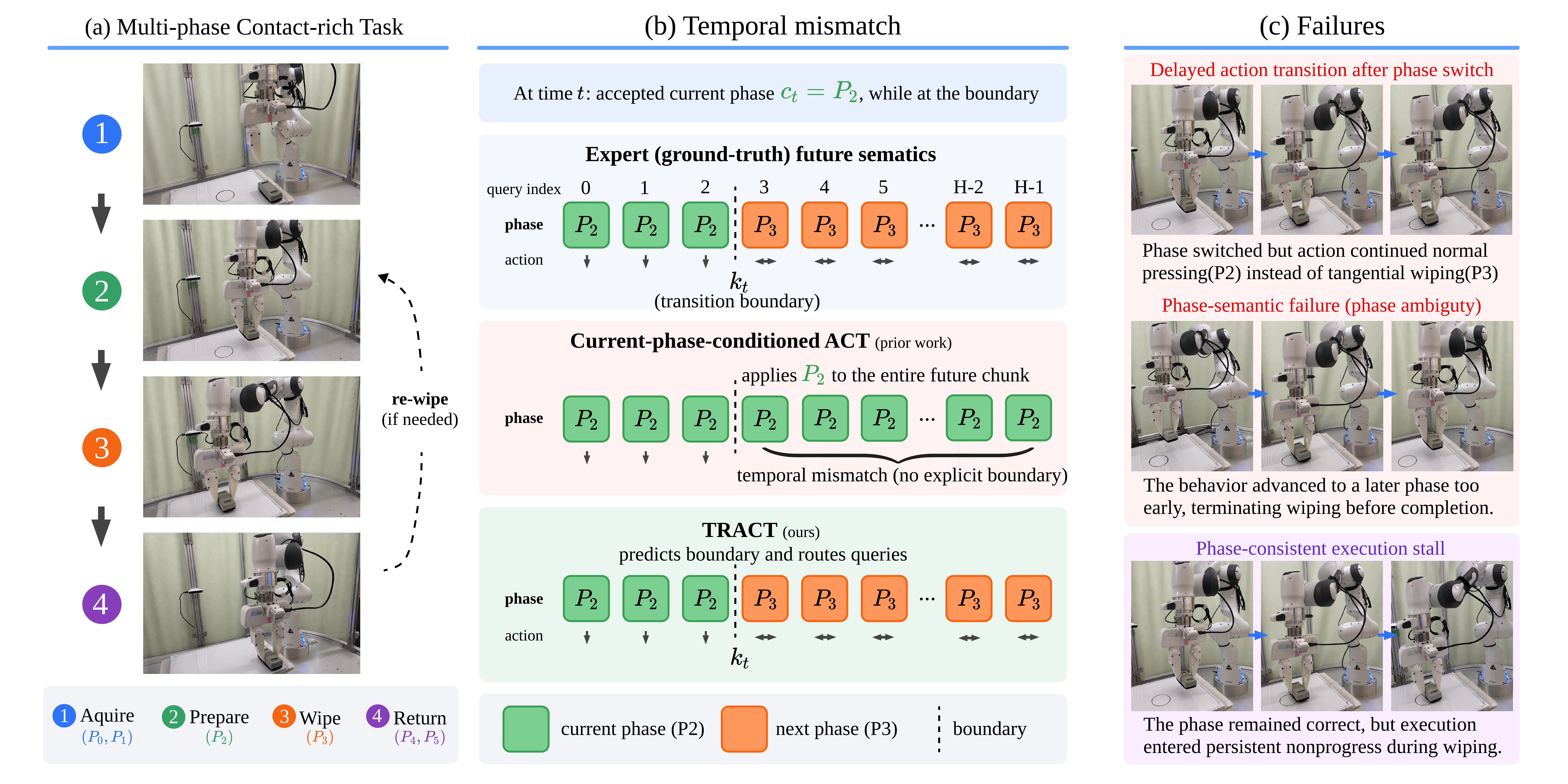}
\refstepcounter{figure}%
\label{fig:motivation}
\parbox{0.965\textwidth}{\footnotesize Fig.~\thefigure. Task, temporal mismatch, and representative failures. (a) Four semantic steps summarize the six controller phases and the authorized P4$\rightarrow$P2 re-wipe. (b) At the P2$\rightarrow$P3 boundary, applying accepted current phase P2 to the full future chunk leaves its suffix temporally mismatched; TRACT predicts $k_t$ and routes that suffix to P3. (c) Representative outcomes include a delayed action transition after a phase switch, premature phase advance that terminates wiping, and phase-consistent execution stall.}
\end{center}
}

\maketitle

\begin{abstract}
Action chunking predicts multiple future actions, while conventional phase conditioning describes the current control instant. When a chunk crosses a procedural boundary, assigning its whole horizon the current phase creates a temporal mismatch. We present TRACT, which separates accepted current-phase authority from a single CURRENT-to-NEXT boundary that monotonically routes future queries through phase-specific paths. Given phase-aligned intent, a causal response-deficit integrator compensates suppressed motion caused by friction and other unmodeled contact dynamics. Across six real-robot variants with ten trials each, full TRACT achieves 10/10 success, 99.00 [88.75, 100.00]\% median [min, max] wipe completion, zero observed phase ambiguity, and zero stalls. The flat $T_U$ retains integration but lacks routing and obtains 3/10 success, whereas routed $T_R$ disables integration and obtains 6/10. This crossed comparison supports that integration cannot substitute for routed semantics, but does not isolate routing. On the same routed checkpoint, integration raises success from 6/10 to 10/10 and reduces stalls from 4/10 to 0/10; chronological authority reduces ambiguity from 8/10 to 0/10.
\end{abstract}

\section{Introduction}

Long-horizon contact-rich manipulation requires a robot to acquire a tool, establish contact, sustain task motion, and return the tool while preserving both physical contact and procedural order. Similar visual and proprioceptive states can occur during contact establishment, repeated wiping, and release, although their valid next actions differ. The resulting failures can include a delayed action transition after a phase switch, premature phase advance that terminates wiping, or phase-consistent stall under suppressed physical response (Fig.~\ref{fig:motivation}).

Action chunking predicts a sequence of future actions at each policy query, reducing the effective decision length and producing smooth behavior from demonstrations. Action Chunking with Transformers (ACT) is a representative approach, combining sequence prediction with a temporal ensemble of overlapping chunks \cite{zhao2023act}. This future-action representation raises a separate temporal question: which procedural phase should each query in the predicted chunk represent?

Let $o_t$ denote the current observation, $c_t$ its accepted phase, and $H$ the chunk horizon. Current-phase conditioning maps $(o_t,c_t)$ to $\mathbf A_t=\{\mathbf a_{t,j}\}_{j=0}^{H-1}$ and thereby attaches one present-time phase variable to the whole future horizon. Near a procedural boundary, however, the expert phase sequence can be
$c_t,c_t,c_t,n(c_t),n(c_t),\ldots$, rather than a phase-homogeneous repetition of $c_t$. This exposes two distinct ambiguities. \emph{Present-phase ambiguity} concerns which phase is valid at the current physical instant. \emph{Future-boundary ambiguity} concerns whether the predicted chunk crosses into the next phase and where that transition occurs. Current-phase conditioning addresses the former; the latter belongs to the predicted horizon.

We introduce TRACT, a temporally routed action-chunk policy that assigns these decisions to their corresponding time scales. A chronological authority accepts the current phase under a task-local transition graph. A structured estimator then predicts a single boundary $k_t$ and routes future queries monotonically from the accepted current phase to its unique legal successor. Thus, the current phase is a property of the present control state, whereas the phase boundary is a property of the future action horizon. Fig.~\ref{fig:motivation}(b) contrasts this routed horizon with a current-phase-conditioned chunk at the P2$\rightarrow$P3 boundary.

Unknown contact dynamics, including friction, can suppress motion even for a semantically correct chunk. Assuming routing has established phase-aligned intent, TRACT's causal integrator accumulates ACK-eligible response deficit along that intent and decays after recovery. It addresses residual physical stalls, not stale or procedurally invalid semantics, while preserving phase, route, and gripper outputs.

This paper makes three contributions:
\begin{itemize}
  \item We formulate phase-structured action chunking as two temporally distinct decisions: present-phase acceptance and future-boundary routing.
  \item We develop chronological phase authority and a graph-constrained, monotone CURRENT-to-NEXT route that assigns phase semantics to future action queries.
  \item We construct a downstream causal closure for friction- and dynamics-induced stalls; under the current packages and setting, routed $T_R$ without integration obtains 6/10 success versus 3/10 for integrator-enabled flat $T_U$.
\end{itemize}
The evaluation is a mechanism-focused feasibility study on one multi-stage planar-wiping task. Each variant is tested in ten complete trials under matched sensing, control, data, action horizon, and hardware conditions.

\section{Related Work}

\textbf{Action chunks as future representations.}
ACT predicts action chunks and temporally ensembles overlapping predictions \cite{zhao2023act}; Diffusion Policy generates receding-horizon action sequences through conditional diffusion \cite{chi2023diffusion}. Subsequent work improves bimanual interdependence, hierarchical control, test-time decoding, token compression, and parallel action decoding \cite{lee2025interact,zhang2024hirt,liu2025bid,pertsch2025fast,liu2026oat,kim2024openvla,kim2025oft}. These methods strengthen sequence generation and coordination. TRACT adds an ordered procedural route to the future sequence, turning a flat action chunk into a temporally heterogeneous object.

\textbf{Phase and semantic conditioning.}
StageACT conditions low-level behavior on task stage to reduce partial observability and behavior aliasing \cite{lee2025stageact}. PHASER combines current-phase conditioning with multimodal phase estimation and recovery logic \cite{chen2026phaser}. These approaches assign procedural semantics to the current state. Semantic action representations also separate intent from execution detail \cite{huang2026mint}. MoE-ACT supervises phase-specific experts and predicts phase/expert weights for future actions \cite{mazza2026moeact}. TRACT parameterizes future semantics with an accepted current phase, its legal successor, and one boundary variable. The resulting route is an ordered two-segment sequence whose monotonicity follows directly from the representation.

\textbf{Chunk execution and physical response.}
Recent methods modify how a generated chunk is consumed. Real-Time Chunking supports asynchronous generation and inpainting \cite{black2025rtc}; Legato learns chunk continuation \cite{liu2026legato}; BID and Temporal Action Selection improve test-time consistency or choose among historical chunks \cite{liu2025bid,weng2025tas}. PACE and dynamic execution-horizon prediction select how much of a chunk to execute \cite{nie2026pace,zhao2026dehp}, while set-supervised diffusion and neural-ODE imitation improve correction learning or multi-skill dynamics \cite{li2026sdp,zhao2024nodeil}. These mechanisms determine when, how much, or which generated action should be executed. TRACT determines which procedural semantics each future query represents during generation. For physical feedback, Reactive Diffusion Policy couples slow action chunks with fast tactile response \cite{xue2025rdp}. TRACT instead uses the visual--proprioceptive command lineage to estimate intent-aligned response deficit and sustain contact motion.

\section{Method}
\begin{figure*}[!t]
\centering
\includegraphics[width=\textwidth]{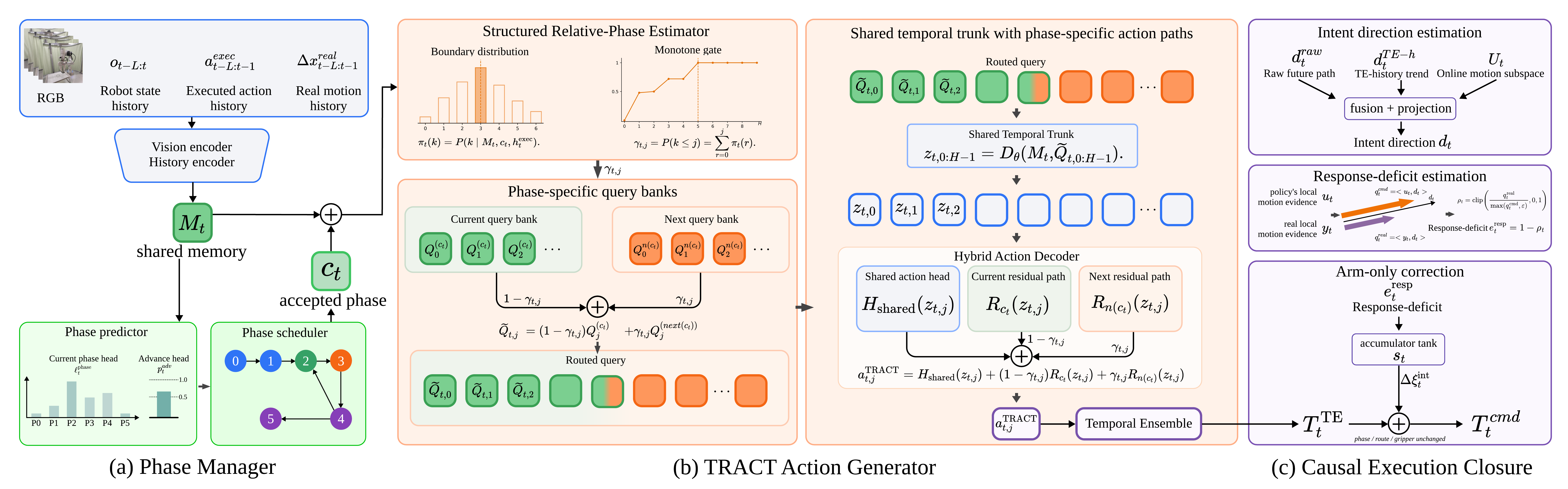}
\caption{Overview of TRACT. (a) The Phase Manager combines a raw phase proposal with dedicated advance evidence and a task-level scheduler to accept the current phase. (b) The action generator represents the future chunk with a structured boundary distribution, a monotone CURRENT-to-NEXT gate, phase-specific query banks, and a shared temporal trunk with phase-specific action paths. (c) The causal execution closure estimates policy intent and ACK-eligible physical response, accumulates persistent response deficit, and applies an arm-only correction while leaving phase, route, and gripper semantics unchanged.}
\label{fig:method}
\end{figure*}

\subsection{Problem Formulation}

We consider a phase-structured manipulation task with observation history $o_{\leq t}$ and strictly past executed commands $\mathbf a^{\mathrm{exec}}_{<t}$. TRACT jointly produces the accepted current phase, an intra-chunk boundary, and a future action chunk:
\begin{equation}
\left(c_t,k_t,\mathbf A_t\right)
=\Pi\!\left(o_{\leq t},\mathbf a^{\mathrm{exec}}_{<t}\right),
\qquad
\mathbf A_t=\{\mathbf a_{t,j}\}_{j=0}^{H-1}.
\label{eq:policy}
\end{equation}
Here $c_t$ belongs to the current physical instant, while $k_t\in\{0,\ldots,H\}$ is the first future query that uses successor semantics. For the legal successor $n(c_t)$,
\begin{equation}
c^{\mathrm{route}}_{t+j}=
\begin{cases}
c_t, & j<k_t,\\
n(c_t), & j\geq k_t .
\end{cases}
\label{eq:route}
\end{equation}
The value $k_t=H$ denotes no transition within the chunk.

\subsection{Chronological Phase Authority}

A shared encoder supplies observation memory to the phase predictor, boundary estimator, and action generator. From causal observation memory and strictly past physical motion, the predictor forms a local state $h_t^{\mathrm{loc}}$ and an episode-level state $h_t^{\mathrm{ep}}$. Its phase head produces raw categorical logits
\begin{equation}
\boldsymbol{\ell}_t^{\mathrm{phase}}
=A_{\mathrm{phase}}\!\left(
h_t^{\mathrm{loc}},h_t^{\mathrm{ep}}
\right),
\label{eq:phase-logits}
\end{equation}
whose maximum-probability class defines the diagnostic proposal $\widetilde c_t$.
A separate authority head conditions its transition evidence on the previous accepted phase:
\begin{equation}
p_t^{\mathrm{adv}}
=\sigma\!\left[
A_{\mathrm{auth}}\!\left(
h_t^{\mathrm{loc}},h_t^{\mathrm{ep}},
\operatorname{Emb}(c_{t-1})
\right)\right].
\label{eq:advance-posterior}
\end{equation}
A deterministic Phase Manager combines $p_t^{\mathrm{adv}}$, explicit rework evidence $e_t^{\mathrm{rw}}$, and valid-tick evidence $v_t^{\mathrm{tick}}$ to accept the phase:
\begin{equation}
\begin{aligned}
c_t&=\mathcal M\!\left(
\widetilde c_t,p_t^{\mathrm{adv}},e_t^{\mathrm{rw}},
c_{t-1},v_t^{\mathrm{tick}}\right),\\
c_t&\in\mathcal L(c_{t-1})
=\{c_{t-1},n(c_{t-1})\}\cup\mathcal R(c_{t-1}).
\end{aligned}
\label{eq:authority}
\end{equation}
where $\mathcal R$ contains explicitly registered task-level rework transitions. The studied task has a P3 self-loop for continued wiping and one P4$\rightarrow$P2 re-wipe edge. Acceptance combines advance/rework evidence with valid-tick and debounce conditions. Each accepted transition atomically updates $c_t$, clears incompatible chunk and temporal-ensemble lineage, and generates a fresh chunk under the new phase. Consequently, P4$\rightarrow$P2 is a present-time authority event; the fresh P2 chunk again uses \eqref{eq:route} for its intra-chunk P2$\rightarrow$P3 boundary.

\subsection{Monotone Intra-Chunk Routing}

Conditioned on accepted $c_t$, the relative-phase estimator predicts a categorical boundary distribution. Its cumulative probability gives the NEXT gate:
\begin{equation}
\begin{aligned}
\pi_t(k)&=P(k_t=k\mid M_t,c_t,h_t^{\mathrm{exec}}),
&& k\in\{0,\ldots,H\},\\
\gamma_{t,j}&=P(k_t\leq j)=\sum_{r=0}^{j}\pi_t(r),
&& \gamma_{t,j+1}\geq\gamma_{t,j}.
\end{aligned}
\label{eq:boundary}
\end{equation}
For an annotated expert chunk, the hard target $k_t^\star$ is the first valid query whose phase label equals the legal successor $n(c_t)$; if no such query occurs, $k_t^\star=H$. The endpoint $k_t^\star=0$ assigns NEXT semantics to the full predicted horizon, whereas $k_t^\star=H$ represents no transition inside it. Terminal phase P5 therefore fixes $k_t^\star=H$. The registered P4$\rightarrow$P2 re-wipe is excluded from this boundary target: once that present-time event is accepted, TRACT clears the old lineage and constructs a fresh P2 chunk.

Because $\gamma_{t,j}$ is a cumulative distribution, every increase is irreversible with query index. The supported route family is exactly a CURRENT prefix followed by a legal-NEXT suffix; oscillating or unrelated phase sequences have zero representational support. This monotonicity is guaranteed before action decoding and is independent of decoder behavior. For phase-specific query banks $Q^{(p)}$, query $j$ becomes
\begin{equation}
\widetilde Q_{t,j}
=(1-\gamma_{t,j})Q_j^{(c_t)}
+\gamma_{t,j}Q_j^{(n(c_t))}.
\label{eq:routed-query}
\end{equation}
The routed action uses the same monotone gate in both the decoder query and CURRENT/NEXT phase-specific residuals:
\begin{equation}
\begin{aligned}
\mathbf z_{t,0:H-1}
={}&D_\theta(M_t,\widetilde Q_{t,0:H-1}),\\
\mathbf a_{t,j}^{\mathrm{TRACT}}
={}&H_{\mathrm{shared}}(\mathbf z_{t,j})
+(1-\gamma_{t,j})R_{c_t}(\mathbf z_{t,j})\\
&+\gamma_{t,j}R_{n(c_t)}(\mathbf z_{t,j}).
\end{aligned}
\label{eq:routed-action}
\end{equation}
The residual paths provide phase-specific pose, trajectory, and gripper semantics while $H_{\mathrm{shared}}$ preserves the common action representation. The route-disabled variant implements a prior-work-style flat phase-conditioned chunk: it retains the shared temporal trunk and accepted current-phase input, but replaces the phase-specific query and residual paths with one shared query/action path. It therefore changes conditioning only when the accepted present phase changes and provides no explicit boundary for the already predicted future horizon.

For boundary target $y_t(k)$, expert action $\mathbf a^*_{t+j}$, and valid horizon mask $\mathcal V_t$, the routed foundation is trained with
\begin{equation}
\begin{aligned}
\mathcal L_{\mathrm{sb}}
&=-\sum_{k=0}^{H}y_t(k)\log\pi_t(k),\\
\mathcal L_{\mathrm{action}}
&=\frac{1}{|\mathcal V_t|}
\sum_{j\in\mathcal V_t}
\ell(\mathbf a_{t,j}^{\mathrm{TRACT}},\mathbf a^*_{t+j}),\\
\mathcal L_{N0}
&=10\mathcal L_{\mathrm{action}}
+\mathcal L_{\mathrm{sb}}
+5\mathcal L_{\mathrm{route}}\\
&\quad+10\mathcal L_{\mathrm{grip}}
+10\mathcal L_{\mathrm{safe}}
+\mathcal L_{\mathrm{alias}} .
\end{aligned}
\label{eq:supervision}
\end{equation}
Ground-truth routes initially select the phase paths, after which an oracle-to-predicted curriculum introduces soft gates. Here $\mathcal L_{\mathrm{sb}}$, $\mathcal L_{\mathrm{grip}}$, $\mathcal L_{\mathrm{safe}}$, and $\mathcal L_{\mathrm{alias}}$ denote structured-boundary, gripper-route, decoded-safety, and strict-alias supervision, respectively. A matched-time conversion maps the source-grid horizon to the 30\,Hz deployment grid. History refinement adds chunk-7D, overlap, ordinary-TE, and boundary terms; N2 uses a weight-0.2 hard-boundary term, while N3 uses the corresponding soft term.
After N0--N3, the Nominal generator and ordinary TE are frozen before the phase and authority heads in \eqref{eq:phase-logits}--\eqref{eq:advance-posterior} are fitted: supervised Manager fitting precedes offline phase-autoregressive calibration. Nominal and Manager parameters therefore never share an optimizer.

\subsection{Causal Execution Closure}

The ordinary temporal ensemble (TE) fuses predictions for the same physical instant only under matching episode, accepted-phase, transition, timestamp, and freshness lineage. Because the execution closure integrates along the policy direction, it assumes that upstream intra-chunk routing has already supplied phase-aligned action semantics; it is not designed to correct a stale or procedurally invalid direction. Define
$\delta_\sigma(T_a,T_b)=\log(T_a^{-1}T_b)^\vee\oslash\boldsymbol\sigma$,
where $\boldsymbol\sigma$ is the action standard deviation. Direction $\mathbf d_t$ fuses the raw within-chunk path with an exponentially weighted TE trend, handles reversal, and compatibility-projects only translation onto the online real-motion subspace while retaining rotation.

An update requires a historical ACK with eligible response timestamp and matching context
$\kappa_t=(c_t,q_t,e_t^{\mathrm{exec}})$, where $q_t$ is the transition sequence and $e_t^{\mathrm{exec}}$ the execution epoch. Expected and realized increments produce
\begin{equation}
\begin{aligned}
\mathbf u_t&=
\begin{cases}
\delta_\sigma(T_t^{\mathrm{raw},0},T_t^{\mathrm{raw},q'}),&
\text{valid raw path},\\
\mathbf g_t^{\mathrm{TE}},&\text{otherwise},
\end{cases}\\
\mathbf y_t&=\delta_\sigma(T_{t-1}^{\mathrm{real}},T_t^{\mathrm{real}}),\\
\rho_t&=\operatorname{clip}\!\left(
\frac{\langle\mathbf y_t,\mathbf d_t\rangle}
{\max(\langle\mathbf u_t,\mathbf d_t\rangle,\epsilon)},0,1\right),\\
e_t^{\mathrm{resp}}&=1-\rho_t,\\
R_t&=\sum_{b\in\{p,\omega\}}\bar w_{t,b}
\frac{[\langle\mathbf V_{t,b},\hat{\mathbf v}_{t,b}\rangle]_+}
{\theta_{b,c_t}},\\
\widetilde e_t&=\operatorname{clip}\!\left(e_t^{\mathrm{resp}}(1-R_t),0,1\right).
\end{aligned}
\label{eq:response}
\end{equation}
Here $\hat{\mathbf v}_{t,b}$ is a unit block of $\boldsymbol\sigma\odot\mathbf d_t$, $\bar w_{t,b}$ its normalized squared weight, and $\theta_{b,c_t}$ a phase-specific speed threshold. Directional speed recovery therefore suppresses further accumulation without a demonstration-response baseline. At control rate $f$, the response-deficit state uses the rise-time coefficient
\begin{equation}
\alpha=1-\exp\!\left[-\frac{1}{f\tau_{\mathrm{rise}}}\right],
\qquad
s_t=(1-\alpha)s_{t-1}+\alpha\widetilde e_t,
\label{eq:integrator-state}
\end{equation}
where $\tau_{\mathrm{rise}}$ is the compensation rise time and $\alpha$ converts it into a per-tick update at control rate $f$. The deployed state and arm command are
\begin{equation}
\begin{aligned}
\Delta\boldsymbol\xi_t^{\mathrm{int}}
&=s_t
\begin{bmatrix}
g_{p,c_t}\hat{\mathbf v}_{t,p}\\
g_{\omega,c_t}\hat{\mathbf v}_{t,\omega}
\end{bmatrix},\\
T_t^{\mathrm{cmd}}&=T_t^{\mathrm{TE}}
\operatorname{Exp}\!\left(\Delta\boldsymbol\xi_t^{\mathrm{int}}\right).
\end{aligned}
\label{eq:integrator}
\end{equation}
The phase-specific $g_{p,c_t},g_{\omega,c_t}$ are translation and rotation growth rates. Confirmed recovery further decays $s_t$ by $\exp[-1/(f\tau_{\mathrm{decay}})]$; invalid lineage, insufficient intent, context change, or inactive phase resets the causal state. The body-frame correction writes only the arm attractor, leaving accepted phase, route, and gripper unchanged.

\begin{table*}[!t]
\caption{Controlled real-robot results. Each variant has ten complete trials. Wipe completion is median [minimum, maximum] over all trials.}
\label{tab:results}
\centering
\begin{tabular}{@{}lcccc@{}}
\toprule
Method & Full-sequence success $\uparrow$ & Wipe completion (\%) $\uparrow$ & Observed phase ambiguity $\downarrow$ & Stall $\downarrow$\\
\midrule
$A_0$: ACT & 0/10 & 0 [0, 0] & N/D$^\dagger$ & 10/10\\
$A_1$: ACT + current phase & 0/10 & 0 [0, 0] & N/D$^\dagger$ & 10/10\\
$T_P$: raw phase switching & 2/10 & 38.63 [24.45, 95.63] & 8/10 & 0/10\\
$T_U$: flat chunk, integrator on & 3/10 & 8.03 [0, 98.45] & 0/10 & 7/10\\
$T_R$: TRACT, integrator off & 6/10 & 77.08 [64.44, 98.60] & 0/10 & 4/10\\
$T_F$: full TRACT & \textbf{10/10} & \textbf{99.00 [88.75, 100.00]} & \textbf{0/10} & \textbf{0/10}\\
\bottomrule
\multicolumn{5}{@{}p{0.96\textwidth}@{}}{$^\dagger$Trials ending in P1 stall did so before an explicit cross-phase action pattern became observable; phase-ambiguity attribution was left unassigned.}\\
\end{tabular}
\end{table*}

\section{Experiments}

\subsection{Task, Data, and Protocol}

We evaluate one fixed-scene, multi-stage planar-wiping task using a Franka Panda, Franka Hand, and Cartesian impedance control. Multi-view RGB and a 13-D Cartesian robot state form the observation. Cameras acquire at 60\,Hz; policy inference and action execution run at 30\,Hz; the low-level controller runs at 1000\,Hz. The action horizon contains 30 queries, approximately one second. All learned components use the same 50 demonstrations. Training follows the separated curriculum above: routed Nominal training and causal-history refinement precede freezing of the action generator and ordinary TE; supervised Manager fitting and offline phase autoregression are then performed without updating the Nominal path. Integrator action normalization and task-specific phase parameters are estimated only after this freeze.

Each variant receives ten complete trials; this is descriptive feasibility evidence for one fixed task. Data, hardware, sensing, action interface, horizon, and ordinary TE are matched. $A_0$ is ACT; $A_1$ adds accepted current phase; and $T_P$ replaces chronological authority with raw phase switching. Flat $T_U$ retains accepted phase, the shared trunk, and the Integrator, but replaces routed paths with a separately trained shared query/action path. Routed $T_R$ disables the Integrator, while $T_F$ enables it on the same $T_R$ checkpoint.

$T_U$--$T_R$ is therefore a crossed comparison of separately trained equal-capacity generators: integration without routing versus routing without integration. It tests whether integration compensates for absent routing, but does not isolate routing. $T_P$--$T_R$ changes runtime authority on a frozen routed policy; $T_R$--$T_F$ changes only the Integrator switch on the same checkpoint.

\subsection{Metrics}

\begin{figure}[!t]
\centering
\includegraphics[width=\columnwidth]{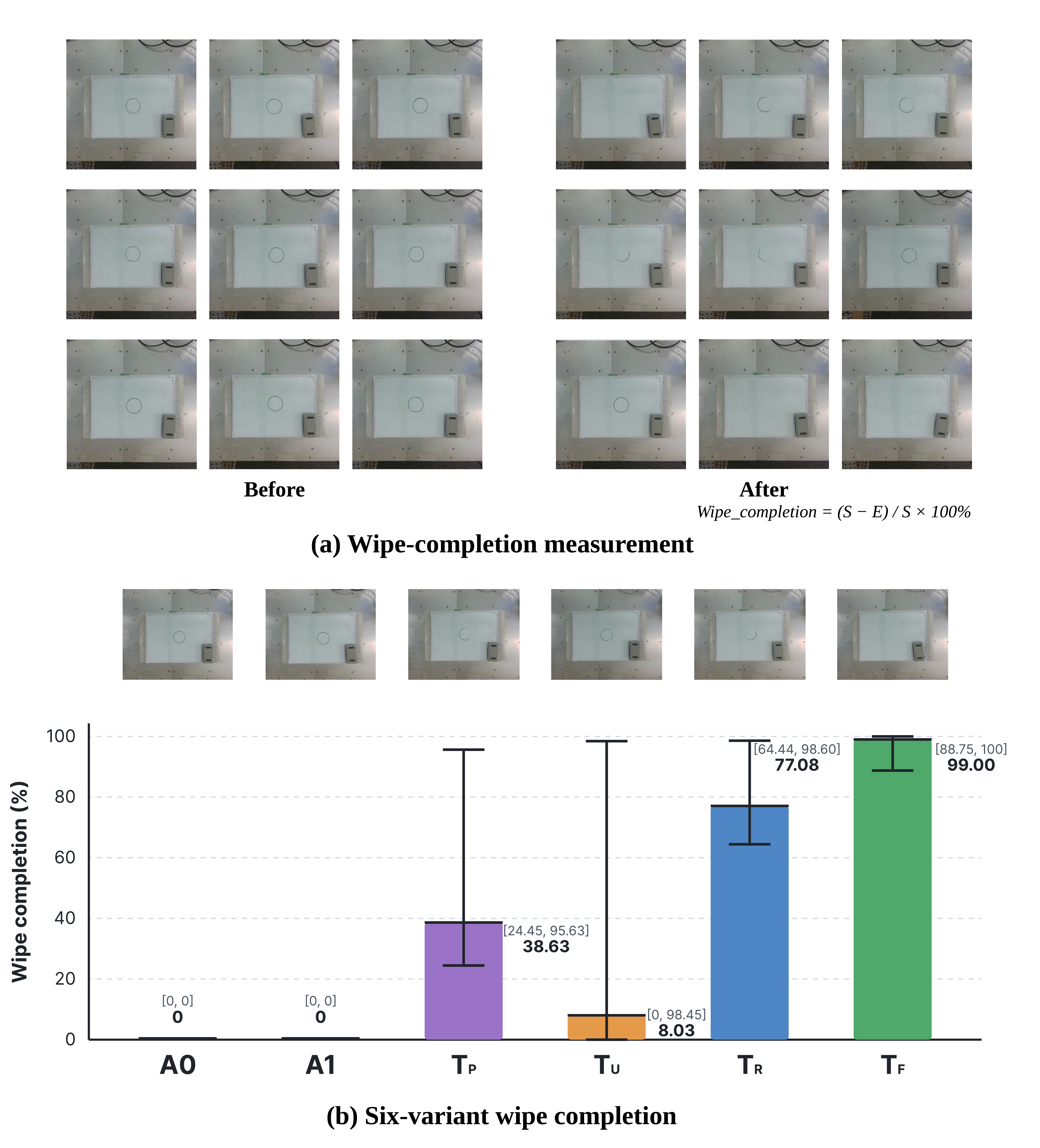}
\caption{Quantitative wiping outcomes over ten real-robot trials per variant. (a) Wipe completion is computed from the removed marker-pixel area between aligned start and end board images. (b) Bars report the median and whiskers indicate the full [minimum, maximum] range. The aligned thumbnails show representative outcomes; aggregate success and failure events are reported in Table~\ref{tab:results}.}
\label{fig:wiping}
\end{figure}

\emph{Full-sequence success} requires autonomous completion in a valid task order, including any P3 continuation or authorized P4$\rightarrow$P2 re-wipe. \emph{Wipe completion} is the removed fraction of central-board marker pixels between global start and end images. Local black-hat enhancement followed by Otsu thresholding yields start and end areas $S$ and $E$; the score is $(S-E)/S\times100\%$. We report median [minimum, maximum] over all ten trials. \emph{Observed phase ambiguity} is a trial-level external review of whether an action semantic pattern is incompatible with the physical procedural context; it does not inspect an internal phase output and therefore also applies to $A_0$. \emph{Stall} denotes persistent valid commands without task progress or autonomous recovery, eventually requiring intervention or termination. The same predeclared intervention criterion is applied to all variants.

\subsection{Results}

\textbf{Full system.}
$T_F$ completes all 10 trials with 99.00 [88.75, 100.00]\% wipe completion, no observed phase ambiguity, and no stall (Table~\ref{tab:results}). Fig.~\ref{fig:wiping} provides the corresponding image evidence and completion distribution.

\textbf{Intra-chunk routing.}
Despite retaining the Integrator, flat $T_U$ obtains 3/10 success, 8.03 [0, 98.45]\% wipe completion, and 7/10 stalls; routed $T_R$ disables it yet reaches 6/10, 77.08 [64.44, 98.60]\%, and 4/10. In representative $T_U$ trials, stale descent actions after P1$\rightarrow$P2 or P2$\rightarrow$P3 disturb contact geometry. Compensation along stale intent cannot repair this semantic mismatch. Thus, under the current packages and setting, integration cannot substitute for routed semantics. Because generator and Integrator both differ, the comparison does not isolate routing. $A_0$ and $A_1$ both obtain 0/10, further showing that unstructured chunks or present-phase input alone are insufficient here.

\textbf{Chronological authority.}
Because $T_P$ removes the advance posterior and chronological Phase Manager, its raw phase switching produces observed phase ambiguity in 8/10 trials, typically a premature P3$\rightarrow$P4 transition soon after wiping begins. Under chronological authority, $T_R$ records 0/10 ambiguity. Its remaining 4/10 failures are phase-consistent mid-wipe stalls, separating procedural acceptance errors from physical response failures. Thus, $T_P$--$T_R$ evaluates chronological phase authority.

\textbf{Causal response integration.}
$T_R$ and $T_F$ share the same routed checkpoint and differ only by the Integrator switch, thereby evaluating integration after phase-aligned semantics are present. Enabling it raises success from 6/10 to 10/10, median wipe completion from 77.08\% to 99.00\%, and reduces stalls from 4/10 to 0/10 with ambiguity fixed at 0/10. This supports intent-aligned compensation for residual response suppression from friction and other unmodeled contact dynamics, not semantic correction.

\textbf{Mechanism audit.}
Frozen replay contains 7,105 crossing chunks among 66,173 horizon-$30$ samples (10.74\%), with monotone routes and 99.80/83.60\% crossing precision/recall. Manager acceptance raises framewise phase accuracy from 83.83\% to 91.83\%, completes 50/50 legal chains, and produces no illegal transition; these diagnostics are separate from Table~\ref{tab:results}'s closed-loop evidence.

\section{Discussion and Conclusion}

The three components address nested failure modes rather than interchangeable sources of performance gain. The Phase Manager constrains the accepted procedural state, the routed representation assigns phase-aligned semantics across the future action horizon, and the Integrator acts only after phase-consistent commands have been generated but the physical response remains suppressed. Accordingly, the 10/10 result should be attributed to the complete stack rather than to either routing or response integration alone. The current evidence isolates the runtime effects of the Phase Manager and Integrator more directly than the independent effect of routing.

Consistently, $T_U$ retains integration without routing (3/10), whereas $T_R$ routes without integration (6/10): under the current packages, integration cannot replace routed semantics, although this crossed comparison does not isolate routing. Once routing is present, same-checkpoint $T_R$--$T_F$ isolates the Integrator's downstream role against residual stalls from friction and other unmodeled contact dynamics, reducing stalls from 4/10 to 0/10 and raising success to 10/10.

The single-boundary contract assumes at most one transition to the selected successor per generated horizon. Horizons spanning two transitions require multiple boundaries, while branching graphs require successor selection. The registered P4$\rightarrow$P2 re-wipe is a new current-time authority event that resets lineage and starts a fresh P2 chunk.

Evidence from one fixed-scene wiping task, one robot, and ten trials per condition establishes feasibility, not cross-task or population-level generalization. $T_U$--$T_R$ is crossed rather than a routing-only intervention, and the Integrator's task- and phase-specific parameters remain untested across motions, contacts, and tasks.

{\renewcommand{\footnotesize}{\fontsize{7.5pt}{8.5pt}\selectfont}
\bibliographystyle{IEEEtran}
\bibliography{references}
}

\end{document}